\documentclass[journal]{IEEEtran}
\pdfoutput=1
\usepackage{cite}
\usepackage{graphicx}
\usepackage{subfigure}
\usepackage{soul}
\usepackage{color} 
\definecolor{yellow}{RGB}{245,225,60}

\ifCLASSINFOpdf
\else
\fi

\begin{document}

\title{Design, Modelling, and Control of a Reconfigurable Rotary Series Elastic Actuator with Nonlinear Stiffness for Assistive Robots}

\author{Yuepeng Qian,
        Shuaishuai Han, 
        Gabriel Aguirre-Ollinger,
        Chenglong Fu
        and Haoyong Yu
\thanks{This work was supported in part by Agency for Science, Technology and Research, Singapore, under the National Robotics Program, with A*star SERC Grant No.: 192 25 00054, and in part by the National Natural Science Foundation of China under Grant U1913205. (Corresponding author: Haoyong Yu.)}
\thanks{Y. Qian is with the Department of Biomedical Engineering, National University of Singapore, Singapore 117583, and also with the Department of Mechanical and Energy Engineering, Southern University of Science and Technology, Shenzhen 518055, China (email: yuepeng.qian@u.nus.edu).}
\thanks{S. Han, G. Aguirre-Ollinger and H. Yu are with the Department of Biomedical Engineering, National University of Singapore, Singapore 117583 (email:han.ss@nus.edu.sg, elgao68@gmail.com, bieyhy@nus.edu.sg).}
\thanks{C. Fu is with the Department of Mechanical and Energy Engineering, Southern University of Science and Technology, Shenzhen 518055, China (email: fucl@sustech.edu.cn).}}


\maketitle

\begin{abstract}
In assistive robots, compliant actuator is a key component in establishing safe and satisfactory physical human-robot interaction (pHRI). The performance of compliant actuators largely depends on the stiffness of the elastic element. Generally, low stiffness is desirable to achieve low impedance, high fidelity of force control and safe pHRI, while high stiffness is required to ensure sufficient force bandwidth and output force. These requirements, however, are contradictory and often vary according to different tasks and conditions. In order to address the contradiction of stiffness selection and improve adaptability to different applications, we develop a reconfigurable rotary series elastic actuator with nonlinear stiffness (RRSEAns) for assistive robots. In this paper, an accurate model of the reconfigurable rotary series elastic element (RSEE) is presented and the adjusting principles are investigated, followed by detailed analysis and experimental validation. The RRSEAns can provide a wide range of stiffness from 0.095 N$\cdot$m/$^\circ$ to 2.33 N$\cdot$m/$^\circ$, and different stiffness profiles can be yielded with respect to different configuration of the reconfigurable RSEE. The overall performance of the RRSEAns is verified by experiments on frequency response, torque control and pHRI, which is adequate for most applications in assistive robots. Specifically, the root-mean-square (RMS) error of the interaction torque results as low as 0.07 N$\cdot$m in transparent/human-in-charge mode, demonstrating the advantages of the RRSEAns in pHRI.
\end{abstract}


\IEEEpeerreviewmaketitle

\section{Introduction}

In recent years, various assistive robots have been developed to physically assist disabled people \cite{aguirre2020lower,chen2019TROelbow,zhong2021toward} or to augment human power \cite{lim2015development}. These robotic devices, such as powered exoskeletons for walking assistance, significantly improve the mobility of people with disabilities and the quality of their lives. In such applications, assistive robots necessarily have direct physical interaction with human \cite{yu2015human}. To improve the performance and safety of physical human-robot interaction (pHRI), the design of the actuator and the corresponding controller are of great importance. In this context, the use of compliance, including active compliance generated by control \cite{duschau2009path} and passive physical compliance \cite{ham2009compliant}, has been considered essential for assistive robots and vital to improve dynamical adaptability and robustness with the environment, and to achieve a safe pHRI \cite{de2008atlas, VIAreview}. Although there have been a number of new developments in the design and control of different compliant actuators, it remains challenging to obtain satisfactory pHRI performance in practical applications.

Series elastic actuator (SEA) \cite{pratt1995SEAs} is the most well-known compliant actuator, in which the physical elastic element is intentionally introduced in series between the stiff actuator and the external load. Many different SEAs have been developed for assistive robots to capitalize on the advantages of the SEA, including lower output impedance, good back-drivability, shock tolerance, energy efficiency, smooth and accurate force transmission, and safety for pHRI \cite{pratt1995SEAs,yu2013control,paine2013design}. However, SEAs typically use springs with fixed stiffness as the elastic element in the force transmission, which is the fundamental limitation of conventional SEAs \cite{yu2013control} as the performance of SEAs is highly dependent on the spring constant \cite{robinson2000design}. On the one hand, soft spring produces high force control fidelity, low output impedance, and reduces stiction, but also limits the force range and the force bandwidth. On the other hand, stiff spring increases the force bandwidth, but reduces force fidelity. To achieve the desired force output and sufficient force bandwidth, most existing SEAs \cite{paine2013design,pspring_2012CompactRSEA,zhang2019clutch,zhang2019admittance} use springs with high stiffness, leading to compromised force control performance, low intrinsic compliance and back-drivability.

In order to overcome the fundamental limitation of conventional SEAs, a number of novel compliant actuators have been proposed \cite{chen2019TROelbow,secondary-2011AwAS-II,secondary-groothuis2013variable,secondary-wolf2011dlr,secondary-sun2018novel,wolf2008nextgeneration,vanderborght2009maccepa,kilic2012synthesis, austin2015control, bai2019MMT, 2021TMECH}. Variable stiffness actuator (VSA) is one of the most investigated examples. VSAs are able to adjust their stiffness based on various working principles \cite{VIAreview}. Among these working principles, tuning the elastic element by a secondary motor and a complicated stiffness adjustment mechanism \cite{secondary-2011AwAS-II,secondary-groothuis2013variable,secondary-wolf2011dlr,secondary-sun2018novel} is the most common approach to achieve stiffness variation. As a consequence, these actuators are generally complicated and heavy, which increases the complexity of the control and makes their deployment in assistive robots difficult, especially in wearable assistive robots. Apart from VSAs, the introduction of nonlinear stiffness in SEAs \cite{chen2019TROelbow,wolf2008nextgeneration,vanderborght2009maccepa,kilic2012synthesis, austin2015control,  bai2019MMT,2021TMECH} also provides a promising solution to the limitation of conventional SEAs. But existing designs still show limitations in achieving nonlinear stiffness and improving adaptability to different applications in assistive robots. In \cite{wolf2008nextgeneration,vanderborght2009maccepa,kilic2012synthesis, austin2015control}, nonlinear stiffness behaviors were achieved with specially designed cam shape, leading to a lack of adaptability to different applications. In \cite{chen2019TROelbow}, a novel series elastic element using linear springs to provide the behavior of nonlinear torsional spring was developed. However, the adjustable parameters in the arrangement of linear springs were not investigated and analysed in detail. Recently, some novel and reconfigurable designs that are able to generate nonlinear and adjustable stiffness behaviors were introduced in \cite{bai2019MMT,2021TMECH}. But the reconfigurability and adjustable stiffness were achieved based on different and complicated winding methods of the pulley blocks, which results in limited model accuracy because of friction and make it more difficult to achieve satisfactory control performance in the pHRI. Thus new designs are desirable to overcome the aforementioned limitations of existing nonlinear SEAs for improving the performance of pHRI and the adaptability to different applications.

In this work, a reconfigurable rotary series elastic actuator with nonlinear stiffness (RRSEAns) is developed for assistive robots. In the RRSEAns, nonlinear and adjustable stiffness profiles are generated by a novel reconfigurable rotary series elastic element (RSEE), which can help achieve a good balance among low output impedance, high fidelity of force control, large force bandwidth and output force range. Linear tension springs are used as the basic elastic element in the reconfigurable RSEE and the stiffness profile is able to be adjusted by changing the setting of the RSEE. Moreover, a kinematic model based on two different adjusting principles is established, followed by detailed analysis and experimental validation, which clearly reveals the influence of the adjustable parameters on stiffness characteristics and also provides guidance for design and stiffness adjustment.

In addition to the actuator design, the controller design is also important for achieving  satisfactory performance and guaranteed safety of the pHRI. For the controller design, nonlinear stiffness usually makes it more difficult to achieve accurate and stable force control. As an easy-to-apply and robust controller, the cascade PID controller \cite{vallery2008compliant} has been a popular choice for force control of linear SEAs \cite{pratt2004late,tagliamonte2014passivity}. With the cascade PID controller, effective and robust force control can be achieved because the velocity-loop bandwidth of the driven motor is much higher than the force-loop bandwidth of the SEA and the motor dynamics is decoupled from the load side. The robustness of the cascade PID controller to external disturbance has been analyzed in \cite{wang2020pid}. In \cite{vallery2008compliant}, a cascade PID controller was designed for a linear SEA, and stability was analyzed with the theory of passivity. In this work, a cascade PI controller is designed for torque control of the proposed actuator with nonlinear stiffness, and is successfully implemented. Based on the adjustable stiffness profiles, tests of human-robot interaction with highly nonlinear and linear stiffness are performed, clearly demonstrating the advantages of nonlinear stiffness in pHRI.

The rest of this paper is organized as follows. Section II describes the design and modelling of the RRSEAns, followed by detailed analysis and experimental validation. The design and analysis of the torque controller are described in Section III. An experimental evaluation on the physical characteristics, control and pHRI performance of the RRSEAns is presented in Section IV. Lastly, the discussion and conclusion are presented in Section V and Section VI, respectively.

\section{Design and Modelling}

\begin{figure*}
  \centering
  \includegraphics[width=0.9\linewidth]{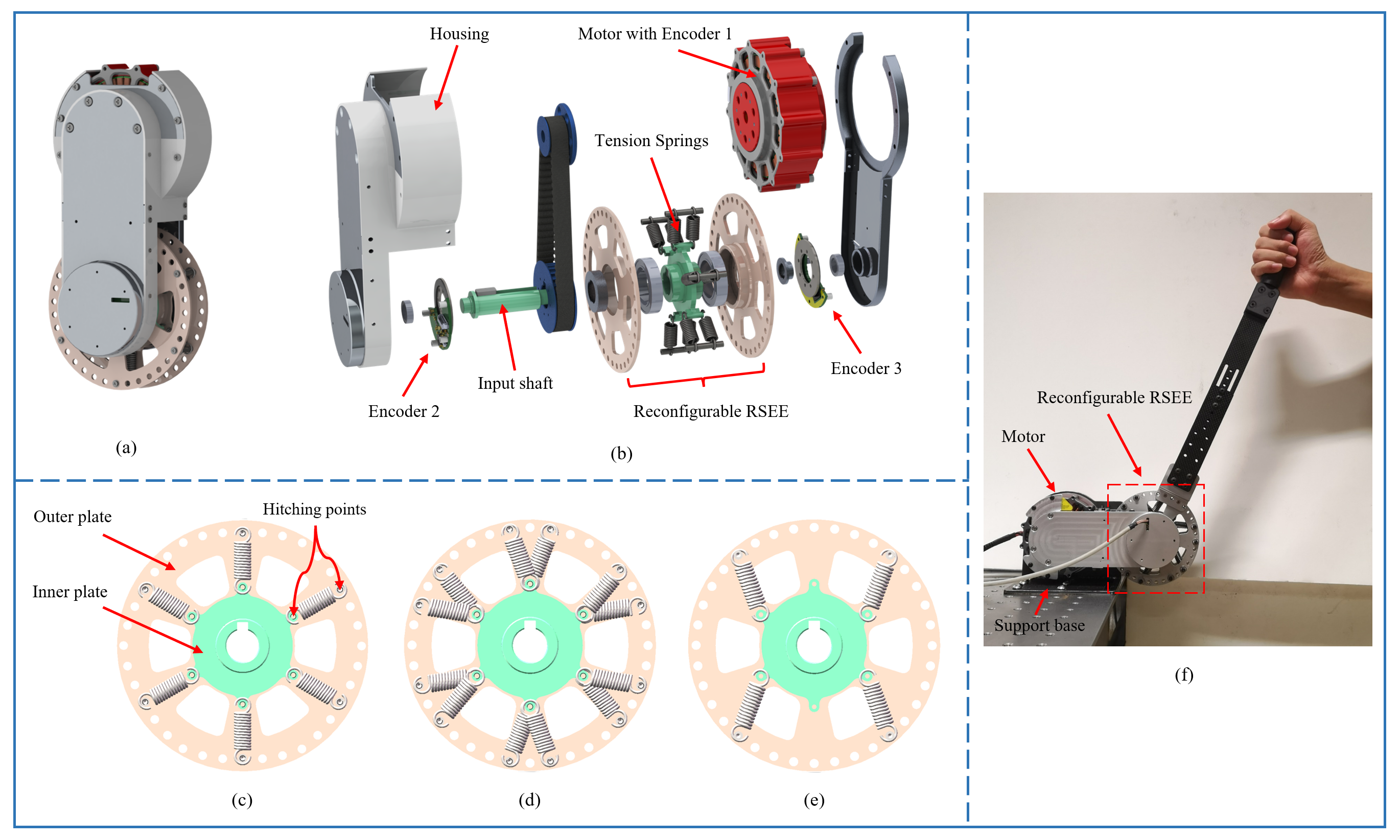}  
  \caption{Mechanical design and prototype of the RRSEAns. (a) Overall view of the CAD model. (b) Exploded view of the CAD model. Three typical configurations of the reconfigurable RSEE are shown: (c) configuration 1 with variable spring pre-tension length (up to 6 spring pairs); (d) configuration 2 with variable offset angle (up to 6 spring pairs); (e) configuration 3 with variable offset angle (up to 4 spring pairs). (f) The RRSEAns prototype.}
  \label{3Dmodel}
\end{figure*}

\subsection{Mechanical Design}

\subsubsection{Actuator Design}

To minimize the size and weight of the RRSEAns, a disk-shaped motor with embedded gear reducer and absolute encoder (QDD-NU80-6, INNFOS Technologies, Beijing, China) was selected to drive the RRSEAns. To further increase the output torque and also reduce the thickness of the RRSEAns, a synchronous belt was used to transmit the force between the motor and the RSEE. The transmission ratio of the synchronous belt is 2:1, which can be further increased to satisfy the requirement for a larger output torque if needed. After the transmission, the RRSEAns can provide a continuous torque of up to 13.2 N$\cdot$m and a peak torque of more than 30 N$\cdot$m, which is adequate for most assistive robots.

Most of the structural parts were made of aluminum alloy. Stress and finite-element analysis simulations were conducted as part of the design and optimization process, which helped to reduce the weight. The total weight of the RRSEAns is approximately 1 kg.

\subsubsection{Reconfigurable Rotary Series Elastic Element}

Series elastic elements can transfer force control to position control due to elasticity, as well as partially masking friction and reflected inertia in motors and transmission mechanisms, allowing accurate torque control and low impedance \cite{pratt1995SEAs,vallery2008compliant}. To eliminate the influence of the transmission error, the deflection angle of the RSEE is measured by two encoders with a resolution of 17 bits (MB039, RLS, Komenda, Slovenia) as illustrated in Fig. \ref{3Dmodel}(b).

In rotary SEAs, the elastic element is used as a torque sensor and a torque generator, so the performance of rotary SEAs largely depends on the characteristics of the elastic element\cite{kong2009control}. Our RSEE features nonlinear stiffness which, compared to linear stiffness, can better meet the requirements of pHRI control. Unlike other existing rotary SEAs that employ custom torsion springs as the elastic element \cite{pspring_2012CompactRSEA,zhang2019clutch,zhang2019admittance,pspring_carpino2012novel,pspring_dos2017design}, a major advantage of the RRSEAns is that the desired nonlinear stiffness characteristic is generated by the novel design of the RSEE with cheap linear springs. The RSEE is designed on the basis of a coaxial rotation mechanism. In this design, two coaxial plates of the RSEE (inner and outer plate), can rotate relatively and are coupled through the tension springs. The inner plate is driven by the motor through the transmission mechanism and the outer plate is linked to the outer load. As a consequence, the compliance from the tension springs is intentionally introduced in series between the input side and the output side of the RSEE. 

As shown in Fig. \ref{3Dmodel}(c), a number of hitching points are placed evenly on the two plates. The tension springs are hitched between the hitching points of the two plates to couple the rotation from the input side to the output side. The stiffness of the coupling is determined by the number of springs, the spring stiffness, the pre-tension length of the spring and the offset angle at the initial position. With this design, variable stiffness values and various stiffness profiles can be obtained by choosing different configuration of the RSEE, which significantly improves the adaptability of the RRSEAns to different applications in assistive robots. Three typical configurations are shown in Fig. \ref{3Dmodel}. The RSEE can support up to 6 pairs of tension springs. The stiffness profiles can be adjusted by choosing different spring pre-tension length in configuration 1, as well as by changing the offset angle at the initial position in configuration 2 or 3.

\subsection{Kinematic Model}

\begin{figure}[tb]
  \centering
  \subfigure[]{\includegraphics[width=0.512\linewidth]{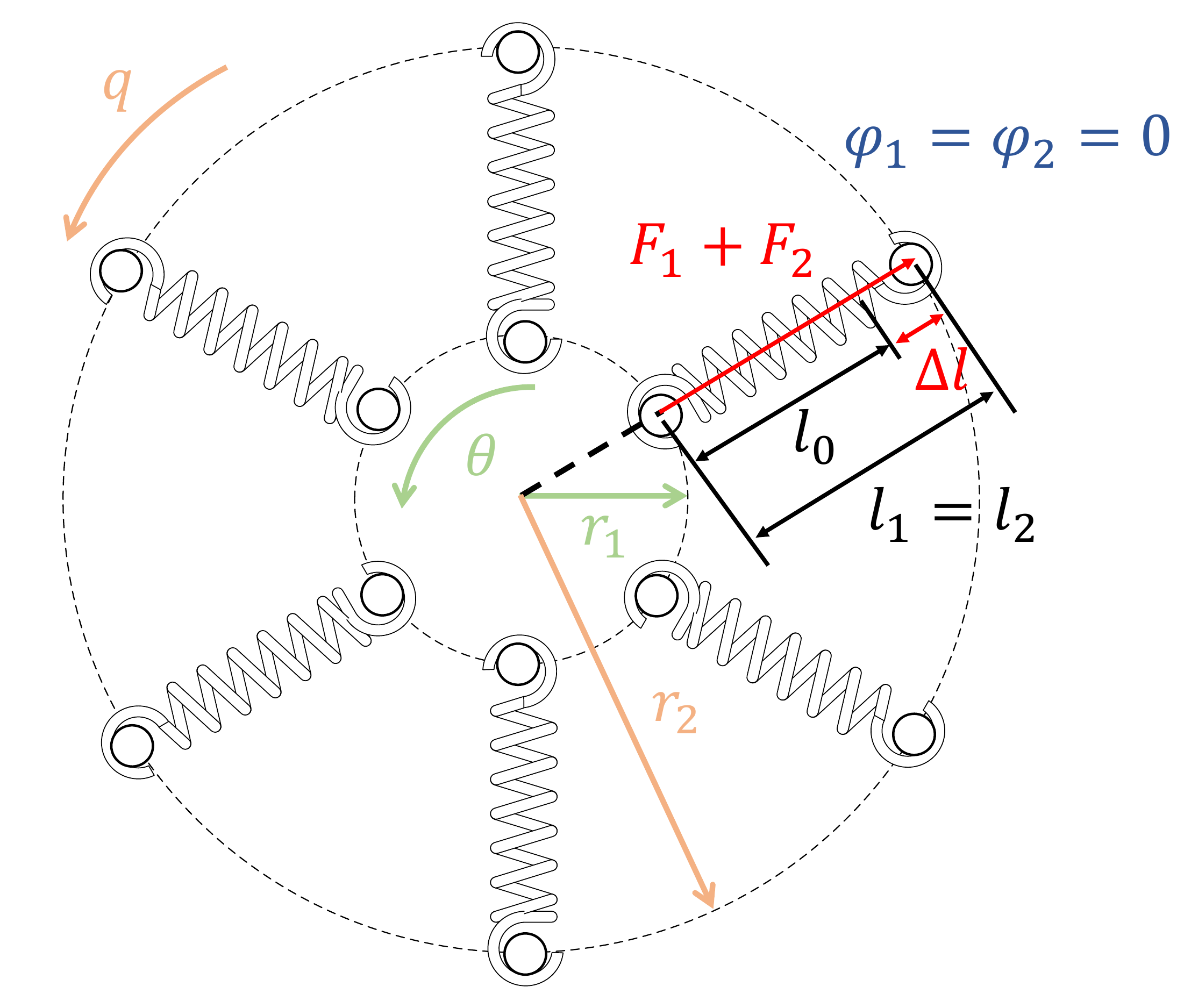}}  
  \centering
  \subfigure[]{\includegraphics[width=0.46\linewidth]{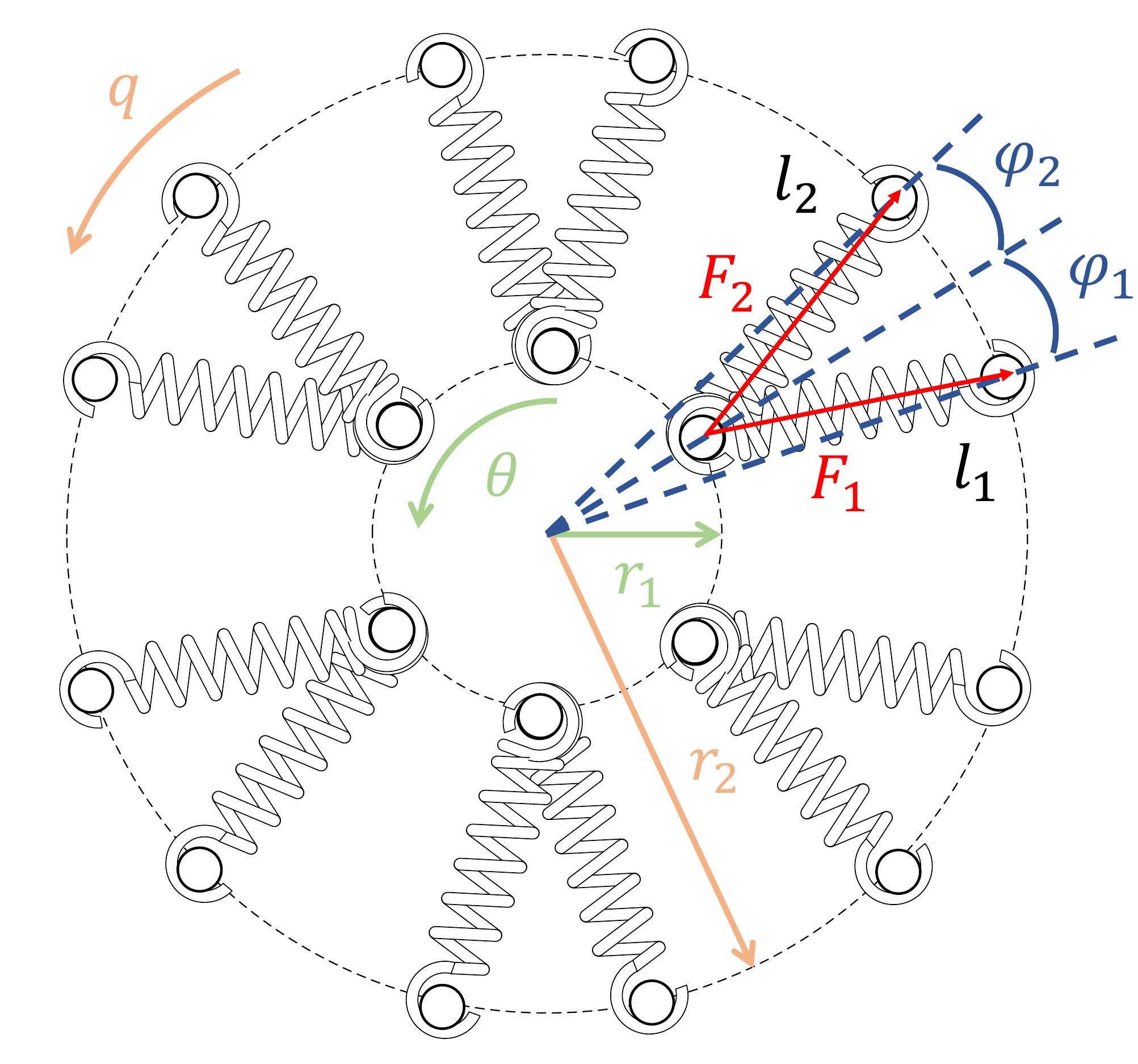}}  
  \caption{Schematic of the reconfigurable RSEE. (a) The configuration with a spring pre-tension. (b) The configuration with an offset angle.}
  \label{ForceAnalysis}
\end{figure}

According to the mechanical design of the reconfigurable RSEE, there are two basic principles of stiffness adjustment: tuning the spring pre-tension length and tuning the offset angle at the initial position, which is described schematically in Fig. \ref{ForceAnalysis}. The geometrical parameters and spring tension force illustrated in Fig. \ref{ForceAnalysis} are defined as:

\begin{itemize}

\item $l_0$, $\Delta l$ denote the spring rest length and the spring pre-tension length, respectively. 

\item $r_1, r_2$ denote the radius of the hitching points of the inner and outer plate, respectively. Specifically, $r_1$ is fixed, and $r_2$ is determined by the spring pre-tension length $\Delta l$ and can be calculated using the following equation:
\begin{equation}\label{r2defination}
{r_2} = {r_1} + {l_0} + {\Delta l}.\\
\end{equation}

\item ${\theta}, {q}$ denote the angle of rotation of the inner and outer plate, respectively.

\item ${\varphi_1, \varphi_2}$ denote the offset angles at the initial position, which are opposite (${\varphi_1}=-{\varphi_2}$) in configuration 2 and identical (${\varphi_1}={\varphi_2}$) in configuration 3.

\item $l_1$ and $l_2$ denote the length of the two springs in each pair at any angle of deflection and can be calculated  based on cosine law using the following equation. Specifically, if the springs is placed with an offset angle (${\varphi_1, \varphi_2}$) at the initial position, the offset angle should be included in the calculation of deflection angle.
\begin{equation}\label{springlength}
\left\{ \begin{array}{l}
{l_ 1 } = \sqrt {r_1^2 + r_2^2 - 2{r_1}{r_2}\cos \left( {{\theta} - {q} + {\varphi_1} } \right)} \\
{l_ 2 } = \sqrt {r_1^2 + r_2^2 - 2{r_1}{r_2}\cos \left( {{\theta} - {q} + {\varphi_2} } \right)} 
\end{array} \right..
\end{equation}

\item $F_i$ (i=1,2) represent the tension force of the two springs in each pair and can be calculated according to Hooke's law as follows:
\begin{equation}\label{springforce}
\begin{array}{c}
{F_{i}} = {k_{s}}\left( {{l_i} - {l_0}} \right).
\end{array}
\end{equation}

\end{itemize}

 With the measurement of ${\theta}$ and ${q}$, the output torque of the RRSEAns can be calculated by:
\begin{equation}\label{Eq_TorqueOutput}
\begin{array}{c}
{\tau _e} = f\left( {{\theta} - {q}} \right) = m{k_{s}}{r_1}{r_2} \Big\lbrack\left( {1 - \frac{{{l_0}}}{{{l_ 1 }}}} \right)\sin \left( {{\theta} - {q} + {\varphi_1} } \right) \\
 + \left( {1 - \frac{{{l_0}}}{{{l_ 2}}}} \right)\sin \left( {{\theta} - {q} + {\varphi_2} } \right)\Big\rbrack 
\end{array}
\end{equation}
where $m$ corresponds to the number of spring pairs. 

The equivalent rotational stiffness of the RRSEAns is defined by:

\begin{equation}\label{Eq_Stiffness1}
\delta{\tau_e}=K_{eq}\cdot \delta\beta
\end{equation}
where $\beta$ is the deflection angle of the RSEE. That is
\begin{equation}\label{Eq_Stiffness2}
K_{eq}=\frac{{\delta\tau _e}}{\delta\beta}=\frac{{\delta\tau _e}}{\delta\left(\theta - q\right)}.
\end{equation}

This kinematic model can represent all configurations of the RSEE. For instance, when the offset angle at the initial position is set to 0 ($\varphi_1=\varphi_2=0$), it characterizes the configuration 1 shown in Fig. \ref{3Dmodel}(c). Also, when the number of spring pairs is set to 4 and the offset angles of the two springs in each pair are set to be identical ($\varphi_1 =  \varphi_2$), it characterizes the configuration 3 shown in Fig. \ref{3Dmodel}(e).

\begin{figure*}
  \centering
  \subfigure[]{\includegraphics[width=0.5\textwidth]{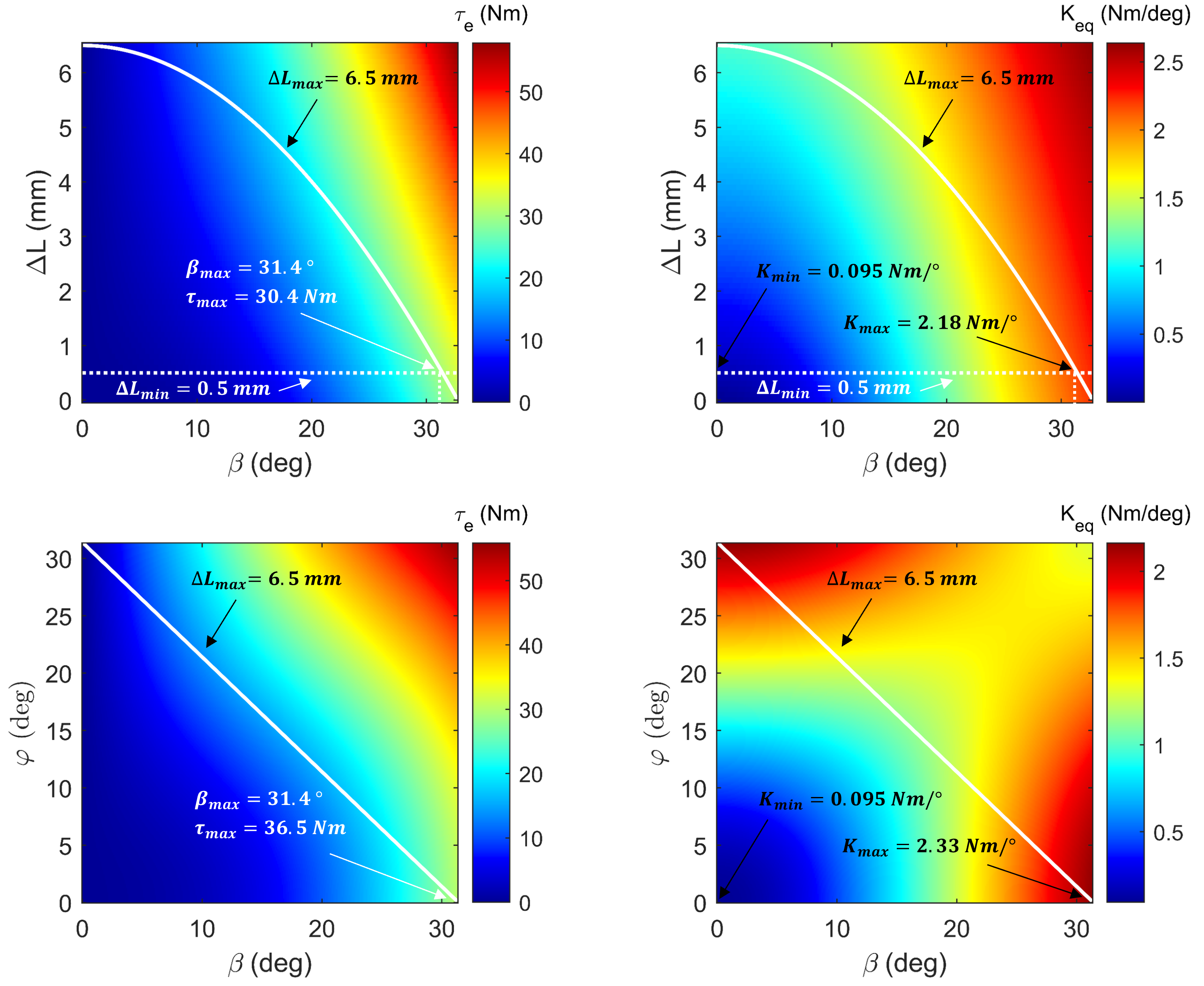}}  
  \centering
  \subfigure[]{\includegraphics[width=0.49\textwidth]{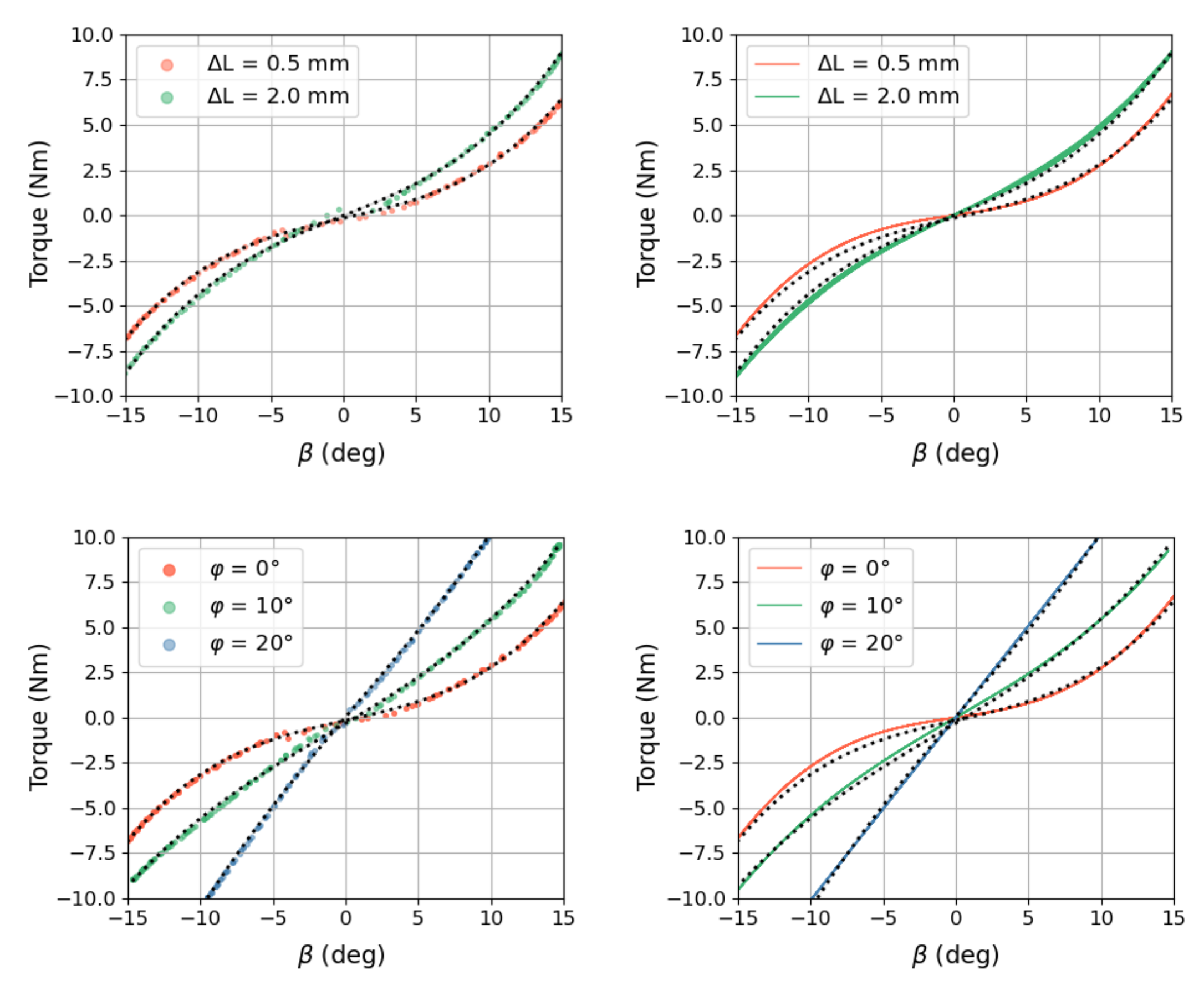}}  
  \caption{(a) Simulation results of the effects of the two adjustable variables on the output torque and stiffness of the RRSEAns. The white solid lines represent the maximum tension length $\triangle L_{max}$ that the selected spring can support, indicating the working space of the RRSEAns. \emph{Top:} Output torque $\tau _e$ v.s. pre-tension length \emph{$\triangle$L} and deflection angle $\beta$;  stiffness $K_{eq}$ v.s. pre-tension length \emph{$\triangle$L} and deflection angle $\beta$. \emph{Bottom:} Output torque $\tau _e$ v.s. offset angle $\varphi$ and deflection angle $\beta$; stiffness $K_{eq}$ v.s. offset angle $\varphi$ and deflection angle $\beta$. (b) Model validation by the experimental results, where the dots, dashed lines and solid lines denote the experimental results, fitting curves of experimental data and simulation results, respectively. \emph{Top:}  The configuration with variable pre-tension length \emph{$\triangle$L}. \emph{Bottom:} The configuration with variable offset angle $\varphi$.}
  \label{stiffness and model}
\end{figure*}

\subsection{Spring Selection}

The number of springs and the spring parameters are determined taking into account the restricted installation space in the compact RSEE, as well as the requirement for a maximum output torque. The RSEE can generate higher stiffness and larger output torque at a large deflection angle with the springs with higher stiffness, while the stiffness of the RSEE around the initial position is still low because of the unique characteristic of the nonlinear stiffness, which can satisfy the need for establishing low impedance in the transparent mode. According to the kinematic model and the geometrical parameters of the RSEE (see Table \ref{paras}), tension springs with a stiffness of 20 kN/m and a rest length of 28.5 mm were selected to meet the design criterion of more than 30 N$\cdot$m torque at approximately 30$^\circ$ of deflection, and preliminary test results showed that the tension springs complied well with Hooke's law.

\subsection{Simulation Analysis}

Based on the kinematic model and the adjusting principles of the RSEE, simulations were performed to reveal the performance limits and characteristics of the RRSEAns. The geometrical parameters of the RSEE and the properties of the selected spring are given in Table \ref{paras}. It is worth noting that in order to overcome the initial tension of the selected spring and avoid the completely relaxed situation of the RSEE ($K_{eq}$ = 0) that may cause some troubles for the use and control of the RRSEAns, the minimum pre-tension length was set to 0.5 mm for all configurations of the RSEE (including configurations with a pre-tension length and configurations with an offset angle). The following specifications were taken into account in the simulations:

\begin{itemize}
\item Stiffness range $K_{eq}$ $\in$ $\left[K_{min},K_{max}\right]$.
\item Maximum allowable output torque $\tau_{max}$.
\item Maximum deflection angle $\beta_{max}$.
\end{itemize}

Fig. \ref{stiffness and model}(a) shows the simulation results of the effects of the two adjustable variables on the output torque and stiffness of the RRSEAns. In the figures, the dashed lines and solid lines represent the minimum pre-tension length and the maximum tension length that the selected spring can support, respectively, which determines the working space of the RRSEAns with any RSEE configuration. For the configuration with a spring pre-tension, the maximum output torque was 30.4 N$\cdot$m while the maximum deflection angle, 31.4$^\circ$, was reached. The stiffness varied from 0.095 N$\cdot$m/$^\circ$ to 2.18 N$\cdot$m/$^\circ$. For the configuration with an offset angle, the output torque reached the maximum, 36.5 N$\cdot$m, with the maximum deflection angle of 31.4$^\circ$. The stiffness can varied from 0.095 N$\cdot$m/$^\circ$ to 2.33 N$\cdot$m/$^\circ$. Based on the simulations, the performance bounds and characteristics were clearly shown. Lastly, and as can be seen in the figures, the performance varied with respect to the RSEE configuration. 

\subsection{Model Validation by Experiments}
A quasi-static test was performed on the bench  test system to validate the accuracy of the kinematic model. During the test, a torque sensor was connected to the RSEE output plate to measure the actual output torque, while the RSEE deflection angle was measured by two absolute encoders. In order to evaluate the torque-deflection characteristics of the RSEE for different configurations, four separate measurements were performed with a pre-tension length of 0.5 mm and 2.0 mm and an offset angle of 10$^\circ$ and 20$^\circ$, as shown in Fig. \ref{stiffness and model}(b). The dots, black dashed lines and solid lines represent the experimental results, the fitting curves and the theoretical results, respectively. The comparison between the experimental and theoretical results found a root-mean-square (RMS) error of 0.28, 0.21, 0.24, 0.22 N$\cdot$m for the four configurations, respectively, which is less than 3.6$\%$, 2.2$\%$, 2.6$\%$, 2.1$\%$ of the peak applied load. The close correlation between the fitting curves of the experimental data and the theoretical predictions shows that the kinematic model is accurate. 

\subsection{RSEE Configuration Analysis}
As shown in Fig. \ref{stiffness and model}, the stiffness performance of the RRSEAns can be significantly adjusted by changing the RSEE configuration. For configuration 1, with the increase of the pre-tension length, the stiffness profile gradually approaches a linear one but it is still nonlinear even with a large pre-tension length because of the design characteristics of the RSEE. For configuration 2, the RSEE with an offset angle is able to vary its stiffness performance in a wider range compared to configuration 1. Hardening, linear and softening modes can be achieved as the offset angle gradually increases, in which the hardening mode is desirable for improving the performance of pHRI and the linear mode can be achieved with an offset angle of approximately 20$^\circ$. Therefore, the offset angle should be set to no more than 20$^\circ$ due to our design target.

Apart from the pre-tension length and the offset angle, the number of springs also has significant influence on the performance of the RSEE. The output torque and stiffness at the same deflection angle are proportional to the number of springs. Therefore, a variety of stiffness and output torque ranges can be achieved by choosing different number of springs. For instance, according to the kinematic model, the output torque and stiffness at the same deflection angle of the configuration 2 is 1.5 times compared to that of the configuration 3 with the same offset angle.

\section{Torque control}

\subsection{Dynamic Model}
\begin{figure}[tb]
  \centering
  \includegraphics[width=1\linewidth]{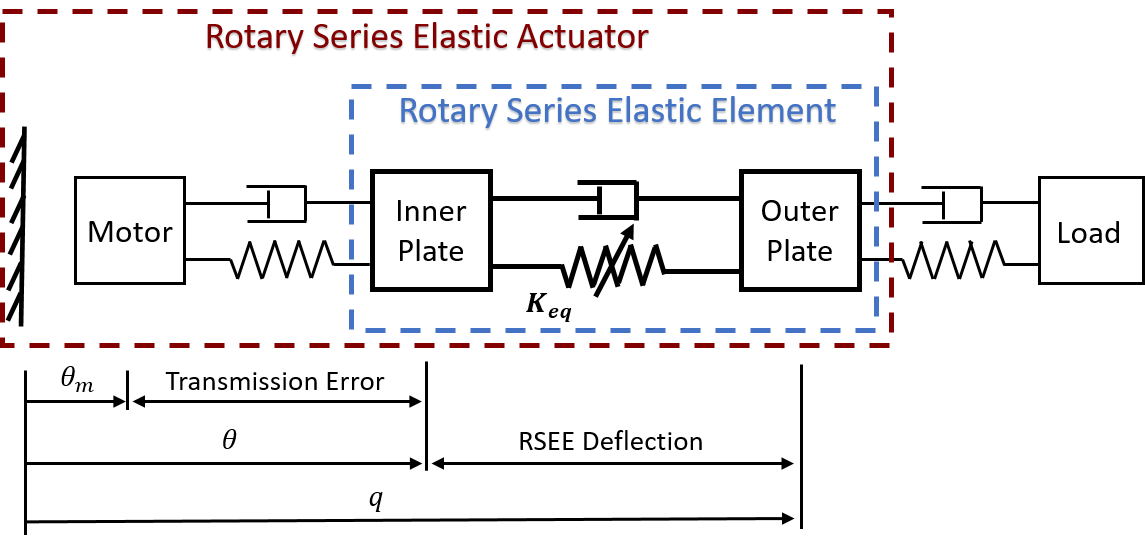}
  \caption{Linearized model of the RRSEAns.}
  \label{Linearized_model}
\end{figure}

The equivalent dynamics from the motor to the output of the RRSEAns can be described as shown in Fig. \ref{Linearized_model}, where the nonlinear stiffness is denoted as $K_{eq}$ in (\ref{Eq_Stiffness2}). The dynamic model of the RRSEAns from the motor current command to the output torque can be formulated as
\begin{equation}
\left\{ \begin{array}{l}
{\tau _{m}} + {J_m}{{\ddot \theta }_m} + {b_m}{{\dot \theta }_m} = {k_m}{i_m}\\
{\tau _e} + {J_s}{{\ddot \theta }} + f\left( {{\theta},{{\dot \theta }}} \right) = 2{\tau _{m}}\\
{\theta _m} = 2{\theta}
\end{array} \right.
\end{equation}
where $\tau_{m}$ is the output torque of the motor, $J_m, b_m, k_m$ are the inertia coefficient, the damping coefficient and the torque constant of the motor, respectively, $i_m$ is the motor current, $J_s$ is the inertia of the reduction gears and the rotating components, and $f\left( {{\theta},{{\dot \theta }}} \right)$ is the friction term. 

By canceling $\tau_{m}$ and $\theta_m$, we can obtain the dynamics of the current command to the output torque of the RRSEAns as 
\begin{equation}\label{Eq_dynamics}
{\tau _e} + \left( {{J_s} + 4{J_m}} \right){\ddot \theta} + 4{b_m}{\dot \theta} + f\left( {{\theta},{{\dot \theta }}} \right) = 2{k_m}{i_m}
\end{equation}
where $\tau _e$ is formulated as equation (\ref{Eq_TorqueOutput}). According to (\ref{Eq_TorqueOutput}) and (\ref{Eq_dynamics}), the nonlinearity of the torque dynamics come mainly from the nonlinear stiffness. 

\subsection{Controller Design and Analysis}

\begin{figure}[tb]
  \centering
  \includegraphics[width=0.48\textwidth]{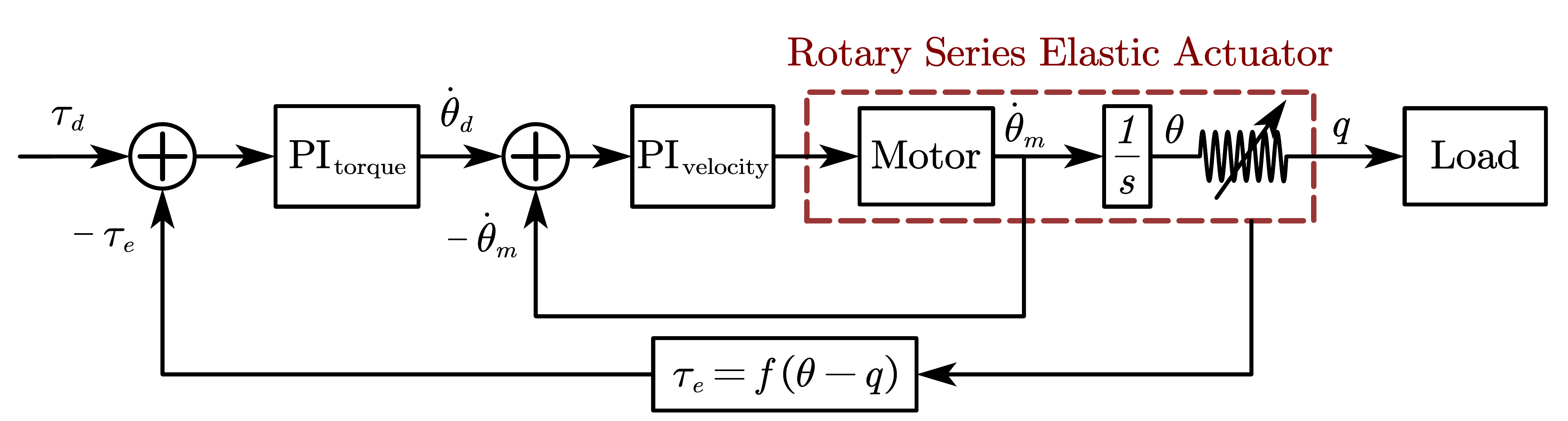}
  \caption{The cascade PI controller.}
  \label{Cascade_PI}
\end{figure}

To perform effective and robust torque control of the RRSEAns, a cascade PI controller was designed and illustrated as in Fig. \ref{Cascade_PI}. The torque control term $PI_{torque}$ is the out-loop controller and generates the command from the inner-loop controller $PI_{velocity}$, according to the torque tracking error. $PI_{velocity}$ should perform fast velocity tracking according to the feedback velocity $\dot \theta_m$ and the desired velocity $\dot \theta_d$. 

The out-loop controller is designed as
\begin{equation}
{\dot \theta _{d}} = \left( {{K_{pt}} + {K_{it}}\frac{1}{s}} \right)\left( {{\tau _d} - {\tau _{e}}} \right)
\end{equation}
where $s$ is the Laplace operator. 

The PI velocity-loop controller of the motor acts as the inner loop, which has a large bandwidth. The velocity controller is designed as
\begin{equation}
{2k_m i_m} = \left( {{K_{pv}} + {K_{iv}}\frac{1}{s}} \right)\left( {{{\dot \theta }_{d}} - \dot \theta } \right)
\end{equation}
To obtain the velocity value $\dot \theta$, a low-pass filter-type differentiator is adopted and designed as
\begin{equation}\label{Differentiator}
    \dot \theta = \frac{\omega_c s}{s + \omega_c}\theta 
\end{equation}
where $\omega_c$ is the cut-off frequency. 

 To verify the effectiveness and robustness of the cascade PI controller, a simulation was performed with MATLAB/Simulink software. The simulation result demonstrated that with proper adjustment of the control parameters ($K_{pv}$, $K_{iv}$, $K_{pt}$, $K_{it}$), accurate force tracking can be achieved. Moreover, since the inner-loop bandwidth is much larger than the outer-loop bandwidth, the cascade PI controller can achieve good robustness to disturbance.
 
 \section{Performance Analysis by Experiments}

Once the RRSEAns was designed and stabilized by applying the control method presented in Section III, the following performance objectives were expected to be verified by experiments:
\begin{itemize}
\item [1)] 
precisely generating the desired torque;  

\item [2)]
generating the desired torque with sufficient bandwidth for human assistance tasks;

\item [3)]
rejecting undesired disturbances, such as rotor inertia, and being robust enough for pHRI;

\item [4)]
exhibiting a low impedance and small interaction force error based on the nonlinear stiffness characteristic;

\item [5)]
being able to adapt to different applications and assistive tasks by means of adjustable stiffness profiles.
\end{itemize}

To clearly compare the effects of nonlinear stiffness on the characteristics and control performance of the RRSEAns, three representative configurations with different levels of nonlinearity were tested and analyzed in detail. The configuration 2 with an offset angle of 0$^\circ$, 10$^\circ$ and 20$^\circ$ were chosen to represent the configuration with high, medium and low nonlinearity, respectively. Details of the differences in stiffness between the chosen configurations can be seen in Fig. \ref{stiffness and model}(b).

\begin{figure}[tb]
  \centering
  \subfigure[]{
  \includegraphics[width=0.41\linewidth]{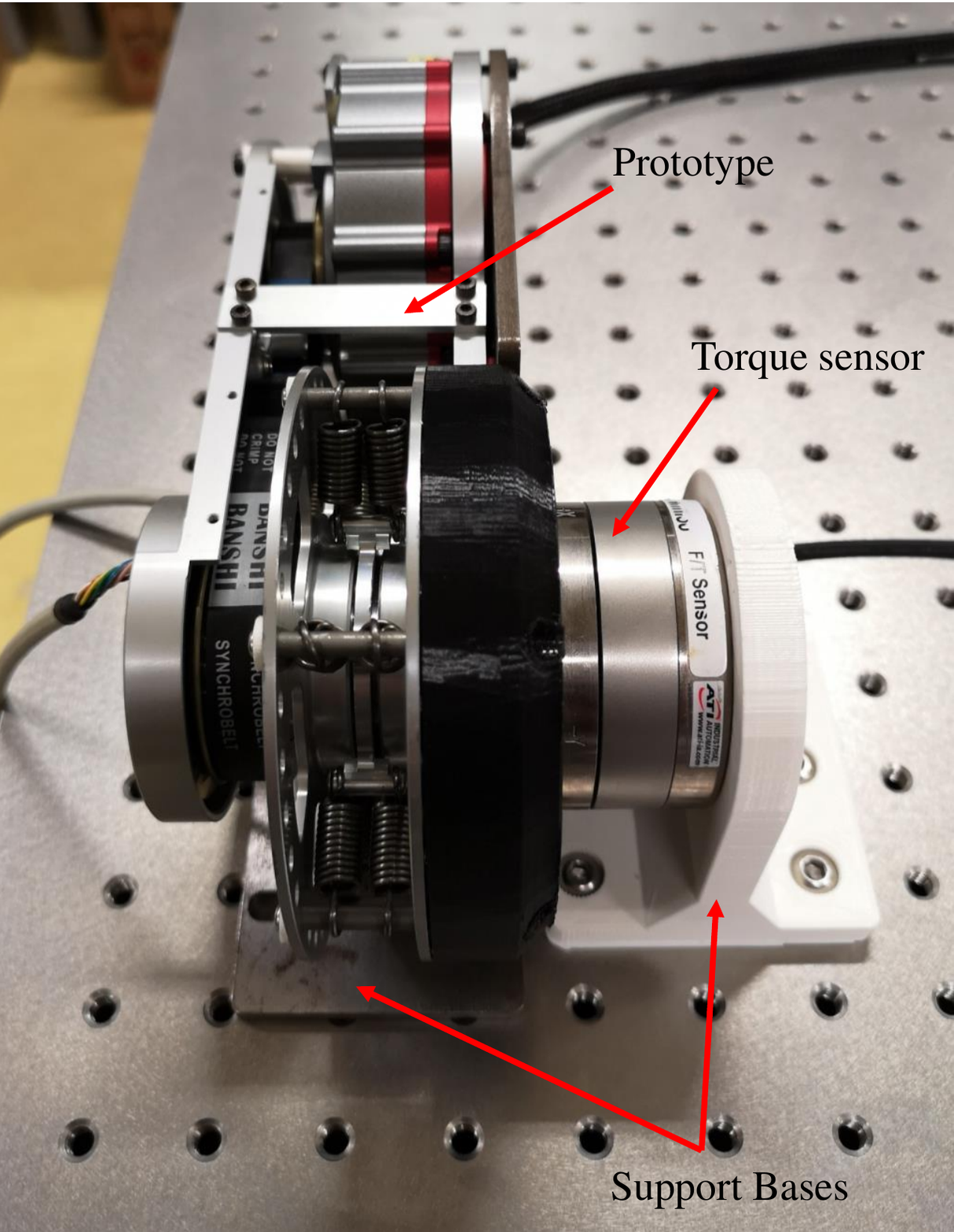}  
  \label{benchtop}}
  \subfigure[]{
  \includegraphics[width=0.486\linewidth]{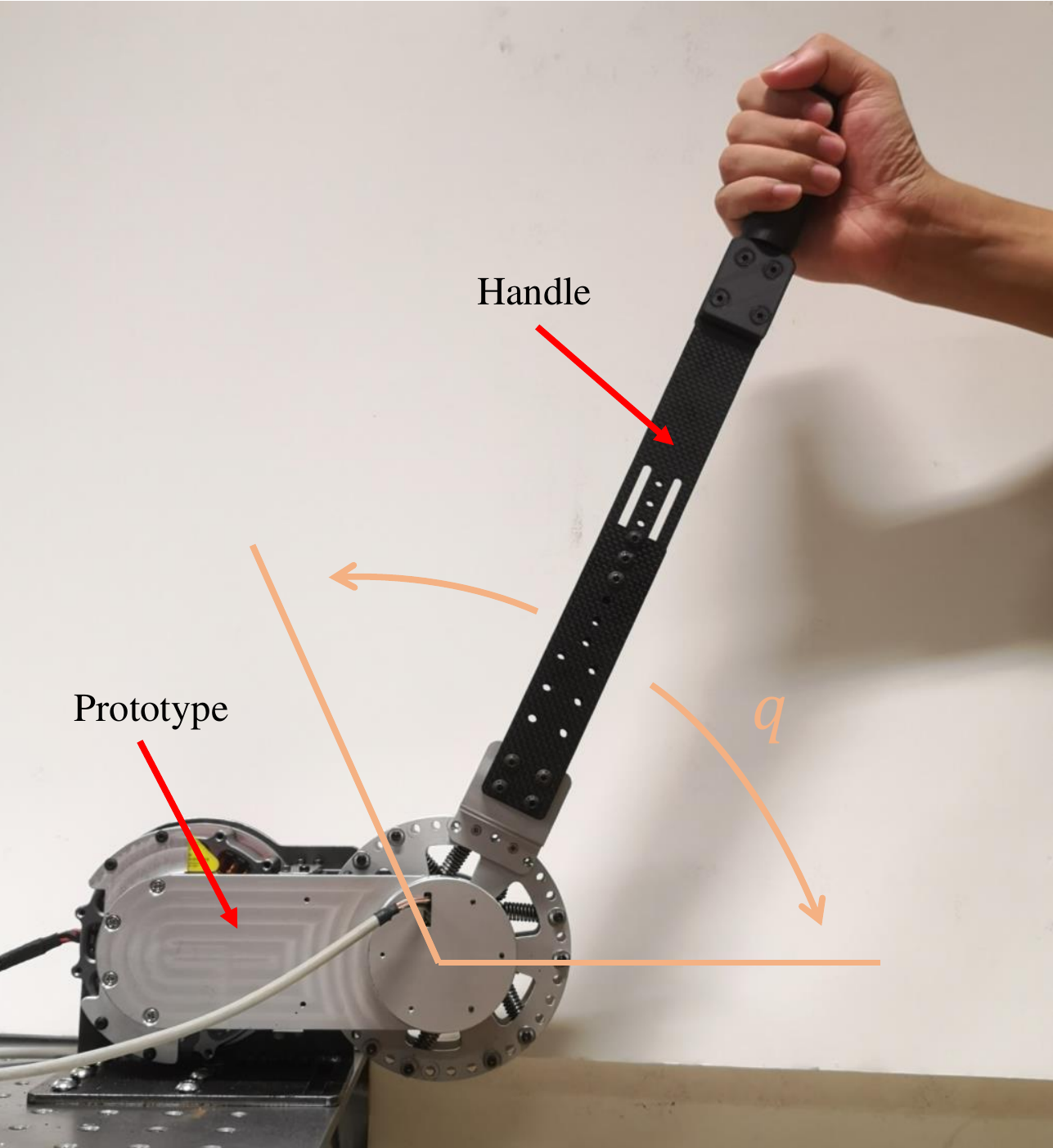}  
  \label{benchtop2}}
  \caption{Experimental setups. (a) The bench test system, consisting of a RRSEAns prototype, support bases, the torque sensor and the control system. (b) The prototype with a handle for the pHRI test.}
\label{benchtop test}
\end{figure}

\begin{table}[tb]
\caption{Parameters of the prototype and controller}
\label{paras}
\begin{center}
\begin{tabular}{l l l}
\hline \textbf{Parameter} &\textbf{Value} & \textbf{Unit}\\
\hline
Size (length $\times$ width $\times$ height) & 194.5 $\times$  103 $\times$ 53.5 & mm\\
Weight & 1.03 & kg\\
Number of springs ($n$) & up to 6 & pair \\
Spring stiffness ($k_{s}$) & 20  & kN/m\\
Spring rest length ($l_{0}$) & 28.5  & mm\\
Radius of inner plate ($r_{1}$) & 24.5 & mm\\
Inertia of structural elements ($J_s$) & 2.33 $\times$ $10^{-5}$ & kg$\cdot$m$^2$ \\
Inertia of the motor ($J_m$)  & 3.06 $\times$ $10^{-4}$ & kg$\cdot$m$^2$\\
Damping of the motor ($b_m$)  & 0 & N$\cdot$ms/rad \\
Torque constant ($k_m$) & 0.44 & N$\cdot$m/A \\
Torque resolution & 0.0064 & N$\cdot$m/deg \\
$K_{pt}$ & 1.5 & -\\
$K_{it}$ & 100 & -\\
$K_{pv}$ & 0.6 & -\\
$K_{iv}$ & 5 & -\\
\hline
\end{tabular}
\end{center}
\end{table}

\begin{figure*}
  \centering
  \includegraphics[width=0.92\linewidth]{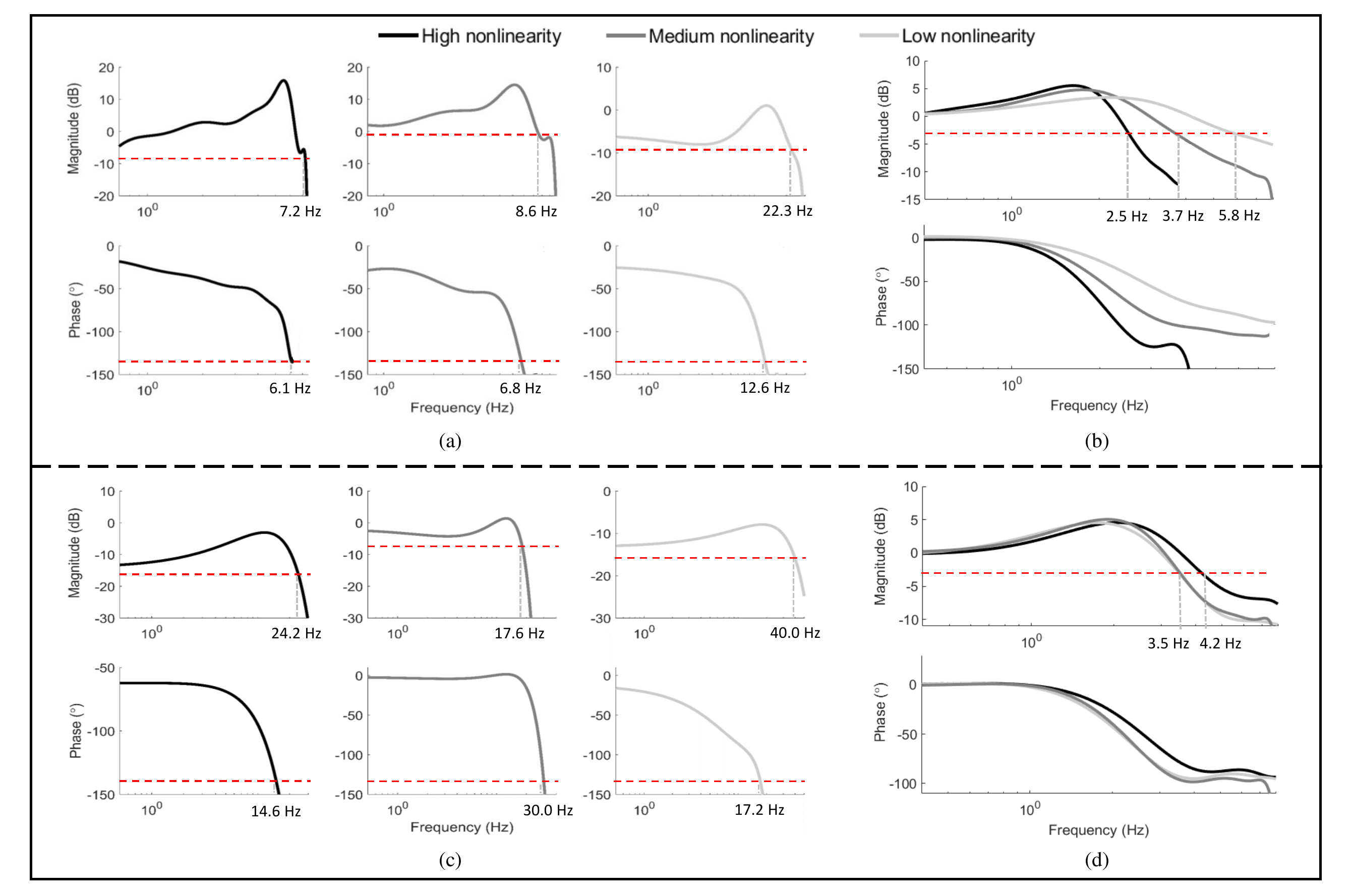}  
  \caption{Frequency responses of the RRSEAns with different level of nonlinearity. (a) The open-loop response with low output torque.  (b) The closed-loop response with low output torque. (c) The open-loop response with high output torque. (d) The closed-loop response with high output torque.}
\label{Bode}
\end{figure*}

\subsection{Experiment Setup}
A prototype of the RRSEAns was built and the design was made according to the CAD model shown in Fig. \ref{3Dmodel}, and the experimental setups are shown in Fig. \ref{benchtop test}. The key parameters of the prototype and the controller are summarized in Table \ref{paras}. Specifically, the controller gains were carefully tuned according to experimental results based on the auto-tuning controller gains yielded with the well-identified torque dynamics and PID Tuner toolbox in MATLAB/Simulink software.

 A control system based on the PC/104 CPU board (PCM-3365N, Advantech, Singapore) was built to control the prototype and collect data from the torque sensor, the absolute encoders and the motor. During the bench test, the prototype was fixed and the outer plate of the RSEE was connected with a six-axis force/torque sensor (Mini58, ATI Industrial Automation, Apex, USA), which is used to measure the actual output torque of the RRSEAns with an accuracy of 1/160 N$\cdot$m. A handle was also connected to the output side of the prototype for the pHRI test.

\subsection{Frequency response}

To verify the physical characteristics of the RRSEAns with different configurations, the open-loop frequency response from the current command to the output torque was tested. Then, with the designed cascade PI controller, the closed-loop frequency response was also tested. Fig. \ref{Bode}(a) and \ref{Bode}(c) show the results of the open-loop response with different levels of nonlinearity. Generally, higher stiffness means higher bandwidth. According to the response with low output torque shown in Fig. \ref{Bode}(a), higher nonlinearity means lower bandwidth, as high nonlinearity corresponds to relatively low stiffness at the same deflection angle as shown in Fig. \ref{stiffness and model}. When the RRSEAns operates with a high output torque, as shown in Fig. \ref{Bode}(c), all bandwidths are much larger due to the relatively high stiffness. Since the minimum open-loop bandwidth is 7.2 Hz and the bandwidth of the human joint is about 4-8 Hz \cite{kong2009control}, the bandwidths of the RRSEAns with different configurations can meet the requirements for human motion assistance. Compared to the open-loop bandwidths, the closed-loop bandwidths shown in Fig. \ref{Bode}(b) and \ref{Bode}(d) are partially decreased. The minimum closed-loop bandwidth is 2.5 Hz, which can still guarantee effective assistance for typical human movements such as walking.

\subsection{Control Performance}

\begin{figure*}
  \centering
  \includegraphics[width=0.96\textwidth]{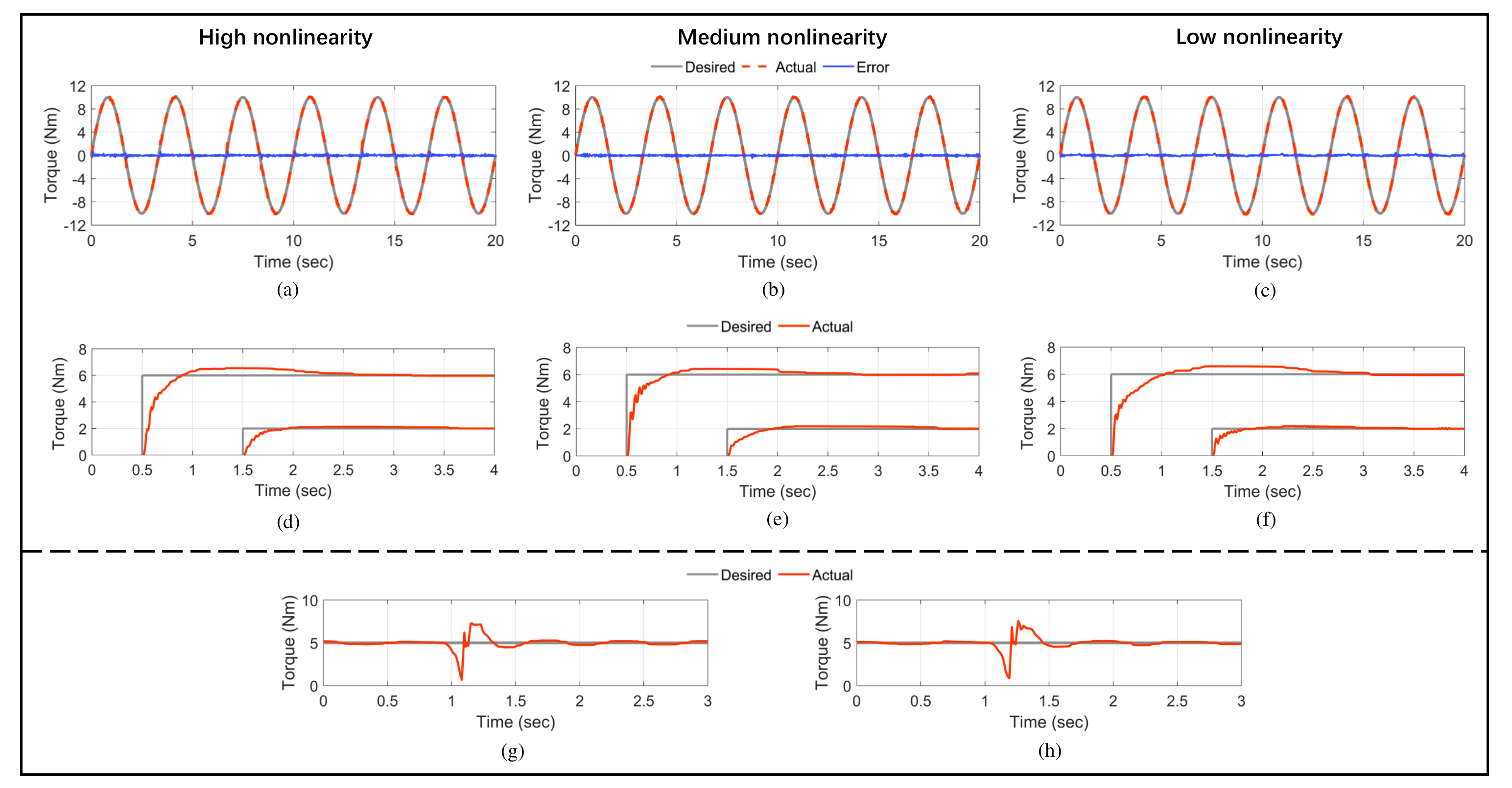}
  \caption{Experimental results of the torque control. \emph{Top:} Torque tracking of a sinusoidal signal with an amplitude of 10 N$\cdot$m and a frequency of 0.3 Hz. \emph{Middle:} Step responses for 2 and 6 N$\cdot$m step inputs. \emph{Bottom:} Collision experiments. (g) The configuration with high nonlinearity. (h) The configuration with low nonlinearity.}
  \label{Torquecontrol}
\end{figure*}

\subsubsection{Torque Tracking and Step Response}
The performance of torque control with the cascade PI controller was evaluated by torque tracking of a sinusoidal trajectory and step response. The torque tracking of a sinusoidal trajectory  with an amplitude of 10 N$\cdot$m and a frequency of 0.3 Hz, and the step responses for 2 and 6 N$\cdot$m step inputs are shown in Fig. \ref{Torquecontrol}. The RMS error of the sinusoidal signal tracking was 0.143, 0.095 and 0.110 N$\cdot$m for the three tested configurations with high, medium and low nonlinearity, respectively. The middle sub-figures in Fig.  \ref{Torquecontrol} show the step response of different configurations. For each configuration, the transient process is fast and satisfactory. According to Fig. \ref{stiffness and model}, the configuration with high nonlinearity means lower stiffness at the same deflection angle compared to the configuration with low nonlinearity, making the torque control more difficult. The torque control results show that with the well-tuned cascade PI controller, the RRSEAns can perform effective and accurate torque tracking with different configurations.

\subsubsection{Robustness Verifying}

This experiment was conducted to test whether the RRSEAns with the cascade PI controller could ensure robust torque control when something unexpected, such as a collision, occurred during pHRI. The collision experiments were performed with both high and low nonlinear configurations. The RRSEAns with different RSEE configurations were controlled with the cascade PI controller without changing the control parameters. During the experiment, a constant torque of 5 N$\cdot$m was commanded, then the handle was released by the operator and the collision occurred. As shown in Fig. \ref{Torquecontrol}(g) and \ref{Torquecontrol}(h), the output torque of the RRSEAns changed abruptly. With the cascade PI controller, the output torque can be quickly recovered to the desired value in a short time interval (about 0.25s) without chattering or tendency to instability. In addition, the good recovery performance from collisions also demonstrates the advantages of the RRSEAns in shock tolerance. 
 
 \begin{figure*}
  \centering
  \includegraphics[width=0.94\textwidth]{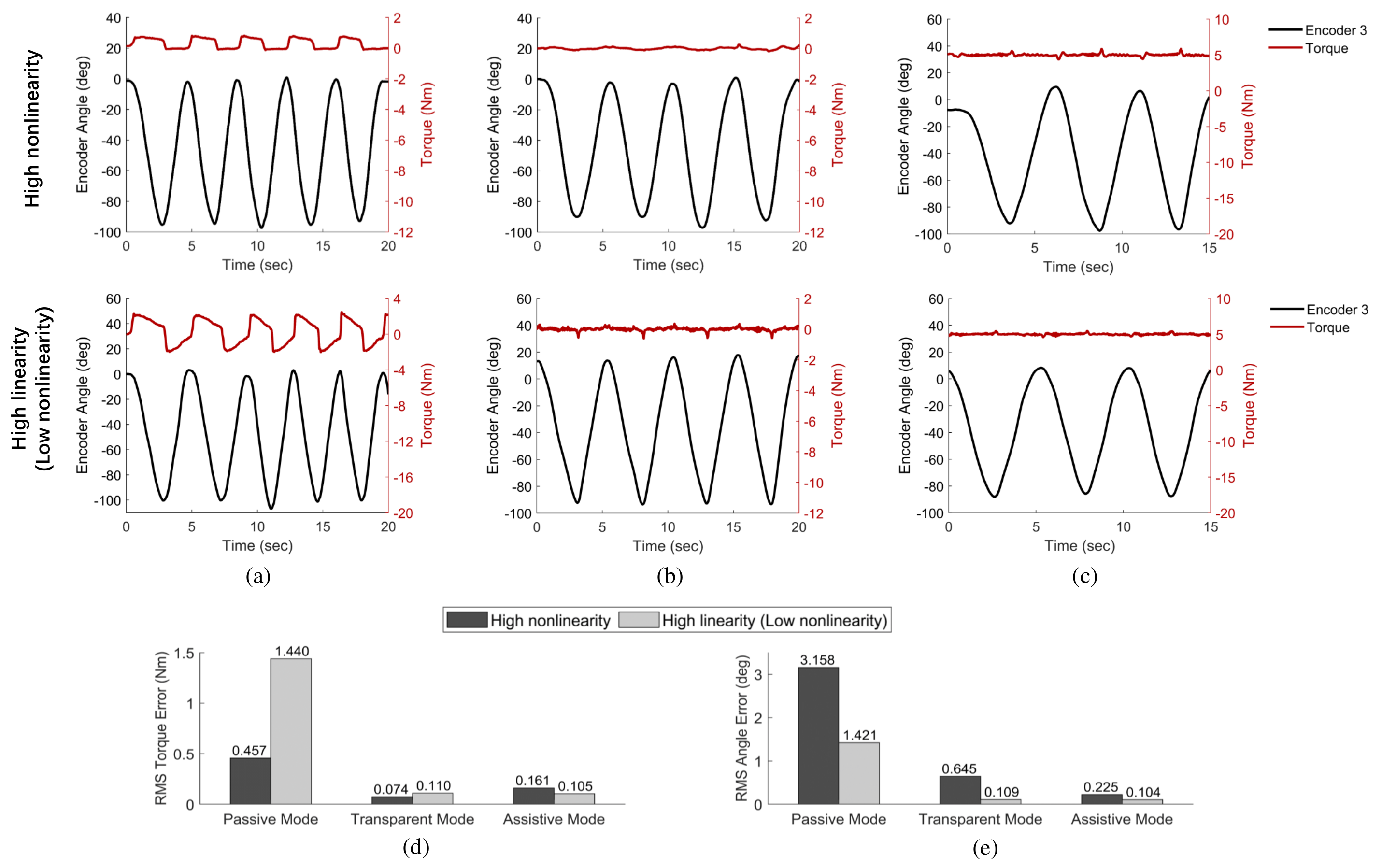}
 \caption{Torque and angle data during the pHRI test, where the handle was moved in a sinusoidal pattern. (a) In the passive mode, the handle was moved without control. (b) In the transparent mode, zero torque was commanded. (c) In the assistive mode, a constant torque (5 N$\cdot$m) was commanded. (d) Comparison of RMS torque errors in each mode. (e) Comparison of RMS angle errors in each mode.}
  \label{pHRI}
\end{figure*}
 
\subsection{Comparison of pHRI Performance}

To verify the pHRI performance of the RRSEAns, three representative working modes (that is, the passive mode, the transparent/human-in-charge mode and the assistive/robot-in-charge mode) were performed. The pHRI test scenario is shown in Fig. \ref{benchtop2}, where the operator interacts with the RRSEAns through the handle. Moreover, in order to compare the characteristics of different configurations and verify the advantages of nonlinear stiffness in the pHRI, the configurations with high nonlinearity and high linearity (low nonlinearity) were tested and the test results are shown in Fig. \ref{pHRI}.

\subsubsection{Passive Mode}
The passive mode means that the RRSEAns was not powered. The handle was moved by the operator without control at approximately 0.25 Hz. The reflective torques and joint angles during the tests were recorded and drawn in Fig. \ref{pHRI}(a). The RMS reflection torque error and the RMS deflection angle error of the configuration with high linearity were 1.440 N$\cdot$m and 1.421$^\circ$. In contrast, the RMS reflection torque error and the RMS deflection angle error of the configuration with high nonlinearity were 0.457 N$\cdot$m and 3.158$^\circ$. By comparing the experimental results, it can be seen that the reflective torque of the configuration with high nonlinearity is lower due to the lower stiffness around the initial position. This means that the configuration with high nonlinearity has a lower mechanical impedance. In addition, the low reflective torques of both configurations demonstrate the high back-drivability of the RRSEAns. 

\subsubsection{Transparent (Human-in-charge) Mode}
The transparent mode is important for the assistance-as-needed (AAN) control of assistive robots, which requires assistive robots to exhibit minimal resistance when the assistance is not necessary\cite{woo2017design}. In this mode, the desired torque is set to be zero to achieve zero impedance control and minimal human interaction force, which also means a human-in-charge mode. Fig. \ref{pHRI}(b) shows the torques and joint angles of the RRSEAns during the transparent mode test. The RMS torque error and the RMS deflection angle error of the configuration with high linearity were 0.110 N$\cdot$m and 0.109$^\circ$ at 0.2 Hz. In contrast, the RMS torque error and the RMS deflection angle error of the configuration with high nonlinearity were 0.074 N$\cdot$m and 0.645$^\circ$ at a similar frequency. The low interaction torque demonstrates the high compliance of the RRSEAns in the transparent mode. Compared to the performance of the configuration with high linearity, the RRSEAns with high nonlinearity can perform smoother and more comfortable transparent movement due to the lower stiffness around the initial position, which also verifies the advantages of nonlinear stiffness in the pHRI.

\subsubsection{Assistive (Robot-in-charge) Mode}
Providing assistance is the fundamental function of the actuators used in assistive robots. The assistive mode test was similar to the transparent mode test, except that a constant assistive torque of 5 N$\cdot$m was commanded and the handle was moved in a sinusoidal pattern at approximately 0.2 Hz. The torques and joint angles of the RRSEAns during the assistive mode tests were recorded and shown in Fig. \ref{pHRI}(c). The RMS torque error and the RMS deflection angle error of the configuration with high linearity were 0.105 N$\cdot$m and 0.104$^\circ$. In contrast, the RMS torque error and the RMS deflection angle error of the configuration with high nonlinearity were 0.161 N$\cdot$m and 0.225$^\circ$. The assistive torque errors of both configurations are very small for torque-based assistive tasks. As compared in Fig. \ref{pHRI}(d) and \ref{pHRI}(e), the RRSEAns with high linearity achieved accurate torque tracking with a lower deflection angle error. It is worth noting that errors in positioning of the RRSEAns results in torque errors that are proportional to stiffness. Therefore, benefiting from the lower stiffness, the RRSEAns with high nonlinearity also performed accurate torque tracking with a higher deflection angle error. The fluctuations in applied torques shown in Fig. \ref{pHRI}(c) are due to relatively slow response of the RRSEAns to the fast changing motion direction. It is difficult to completely remove these errors by increasing the controller gains.

\section{Discussion}

In this work, a reconfigurable rotary series elastic actuator (RRSEAns) for assistive robots was developed. Nonlinear stiffness is used to cancel the limitations of existing compliant actuators and to improve the performance of pHRI. The low stiffness range and high stiffness range of the nonlinear stiffness can provide the soft spring and stiff spring behaviours, respectively, which usually requires two springs with different stiffness \cite{yu2013control} or a complicated stiffness regulating mechanism \cite{secondary-2011AwAS-II,secondary-choi2011robot,secondary-groothuis2013variable,secondary-sun2018novel}. The passive and continuous stiffness change due to the nonlinear stiffness can occur significantly faster than the active stiffness change and also improves the robustness of the system \cite{2015generalizing_torque_control}. In addition, the nonlinear stiffness is generated by a novel RSEE with linear tension springs instead of a custom torsion spring with nonlinear stiffness, which significantly reduces the difficulties for actuator design and modelling.

Compared to existing rotary SEAs that use a high stiffness torsion spring (up to 800 N$\cdot$m/rad) as the elastic element \cite{pspring_2012CompactRSEA,paine2013design,zhang2019clutch,zhang2019admittance}, the stiffness of the RRSEAns (5.4 to 133.5 N$\cdot$m/rad) is much lower, which significantly improves the compliance in the pHRI and, therefore, the user's safety. However, the low stiffness and the nonlinear characteristic often make it more difficult to control the compliant actuator. Compared to other existing rotary SEAs, the RRSEAns controlled by the cascade PI controller has achieved accurate torque tracking with low stiffness. The RMS error of the torque tracking varied from 0.095 to 0.143 N$\cdot$m, depending on the RSEE configuration. In contrast, the SEAC with a high stiffness of 397 N$\cdot$m/rad reported an RMS error of 0.307 N$\cdot$m in the torque tracking test \cite{zhang2019clutch}.

Improving the pHRI performance of assistive robots is an important motivation for us to develop such a rotary SEA with nonlinear stiffness. Test results in the passive mode have shown the low mechanical impedance and high back-drivability of the RRSEAns. In the transparent mode, the RMS of the residual torque was as low as 0.074 N$\cdot$m for the configuration with high nonlinearity, which demonstrates the advantages of the nonlinear stiffness in the pHRI. Compared to the rotary SEA with high linear stiffness (800 N$\cdot$m/rad) \cite{zhang2019admittance}, in the transparent mode, the maximum residual torque of the RRSEAns was only 0.258 N$\cdot$m, which was reported as no more than 1.4 N$\cdot$m in the cited reference. The minimized resistance to human movements in the passive and transparent modes significantly improves the safety of pHRI, which is one of the most important aspects of actuator design and operation due to the close interaction among humans and assistive robots \cite{cordero2014experimental}. Moreover, in the assistive mode, the nonlinear RRSEAns also showed satisfactory torque control performance compared to existing nonlinear SEAs. The RMS torque error of the configuration with high nonlinearity was 0.16 N$\cdot$m, that is, 3.2$\%$ of the applied assistive torque (5 N$\cdot$m). In contrast, the RMS torque error was reported to be 0.27 N$\cdot$m, that is, 4.2$\%$ of the applied assistive torque (6.5 N$\cdot$m) in \cite{chen2019TROelbow}. The accurate torque control guarantees effective assistance for humans.

Adjustable stiffness profiles are another important feature of the RRSEAns, which can be achieved by changing the configuration of the RSEE. Task-specialized optimization can be achieved based on the adjustable stiffness profiles, and different stiffness profiles are suitable for different applications. For example, as compared and shown in Fig. \ref{pHRI}(d) and \ref{pHRI}(e), the configuration with high nonlinearity is more suitable for robots designed for upper limb rehabilitation or other assistive tasks that requires low impedance and accurate control of the interaction force. In the case of lower limb exoskeletons intended to correct abnormal gait, the configuration with medium or low nonlinearity may be a better option, leading to less position error, larger torque output and bandwidth.

Despite the promising results, the proposed RRSEAns and the corresponding controller still have some limitations. The main limitation is the decrease in the closed-loop bandwidth. To ensure stable and robust torque control for different tasks, the control parameters of the cascade PI controller cannot be too large, which limits the controller's bandwidth. The tradeoff between control performance and bandwidth further limits the closed-loop bandwidth of the RRSEAns, which is shown in Fig. \ref{Bode}. In addition, a controller with constant parameters is usually suitable for linear dynamics. Therefore, there is still ample room to make the torque controller adaptive or optimal to be more suitable for the nonlinear RRSEAns.

\section{Conclusion}

In this work, we presented and characterized the RRSEAns, a reconfigurable rotary series elastic actuator designed for assistive robots. The fully assembled RRSEAns weighs only 1.03 kg with an output torque of more than 30 N$\cdot$m, which is suitable for most assistive robots. Unlike conventional SEAs, the RRSEAns features the nonlinear and adjustable stiffness profiles achieved by a novel reconfigurable RSEE with linear tension springs. With an effective and robust cascade PI controller, the performance and advantages of the proposed device (e.g. low mechanical impedance, accuracy and robustness of torque control, and pHRI performance) have been verified by experiments.

In the future, we will seek a better control method for this reconfigurable rotary SEA with nonlinear stiffness to further improve its performance, including but not limited to torque control accuracy, closed-loop bandwidth and assistive control performance. Further engineering development and research are also needed to apply the RRSEAns to actual assistive robots and make the robots useful for rehabilitation or human augmentation applications. 



\bibliographystyle{unsrt}

\bibliography{refer_Exo.bib,refer_SEA.bib,refer_Yu.bib}

\end{document}